%% file: sample-manuscript.tex
\documentclass[manuscript,screen]{acmart}
\AtBeginDocument{%
  }

\usepackage{graphicx}
\usepackage{multirow}
\usepackage{multicol}
\usepackage{tabularx}
\usepackage{booktabs}

\usepackage{amssymb}
\usepackage{pifont}
\usepackage{xcolor}
\usepackage{bbding}
\usepackage{utfsym}
\usepackage{hyperref}
\usepackage{enumitem}
\usepackage{setspace}
\usepackage{courier} % light font weight
\usepackage{url}            % simple URL typesetting
\usepackage{nicefrac}       % compact symbols for 1/2, etc.
\usepackage{microtype}      % microtypography
\usepackage[edges]{forest}
\definecolor{hidden-red}{RGB}{205, 44, 36}
\definecolor{hidden-blue}{RGB}{194,232,247}
\definecolor{hidden-orange}{RGB}{243,202,120}
\definecolor{hidden-green}{RGB}{144,238,144}
\definecolor{hidden-pink}{RGB}{255,245,247}
\definecolor{hidden-black}{RGB}{20,68,106}

\begin{document}

\title{A Survey of Test-Time Compute: From Intuitive Inference to Deliberate Reasoning}

\author{Yixin Ji}
\email{jiyixin169@gmail.com}
%\orcid{0009-0006-9309-3323}
%\email{webmaster@marysville-ohio.com}
\affiliation{%
  \institution{Soochow University}
  \city{Suzhou}
  \country{China}
}

\author{Juntao Li}
\affiliation{%
  \institution{Soochow University}
  \city{Suzhou}
  \country{China}}
\email{ljt@suda.edu.cn}

\author{Yang Xiang}
\affiliation{%
  \institution{Soochow University}
  \city{Suzhou}
  \country{China}}

\author{Hai Ye}
\affiliation{%
  \institution{National University of Singapore}
  \country{Singapore}
}

\author{Kaixin Wu}
\affiliation{%
 \institution{Ant Group}
 \city{Hangzhou}
 \country{China}}

\author{Kai Yao}
\affiliation{%
  \institution{Ant Group}
  \city{Hangzhou}
  \country{China}}

\author{Jia Xu}
\affiliation{%
  \institution{Ant Group}
  \city{Hangzhou}
  \country{China}}

\author{Linjian Mo}
\affiliation{%
  \institution{Ant Group}
  \city{Hangzhou}
  \country{China}}

\author{Min Zhang}
\affiliation{%
  \institution{Soochow University}
  \city{Suzhou}
  \country{China}}
\email{minzhang@suda.edu.cn}

\renewcommand{\shortauthors}{Ji et al.}

\begin{abstract}
The remarkable performance of the o1 model in complex reasoning demonstrates that test-time compute scaling can further unlock the model's potential, enabling powerful System-2 thinking.
However, there is still a lack of comprehensive surveys for test-time compute scaling.
We trace the concept of test-time compute back to System-1 models. 
In System-1 models, test-time compute addresses distribution shifts and improves robustness and generalization through parameter updating, input modification, representation editing, and output calibration. 
In System-2 models, it enhances the model's reasoning ability to solve complex problems through repeated sampling, self-correction, and tree search.
We organize this survey according to the trend of System-1 to System-2 thinking, highlighting the key role of test-time compute in the transition from System-1 models to weak System-2 models, and then to strong System-2 models.
We also point out advanced topics and future directions.\footnote{\url{https://github.com/Dereck0602/Awesome_Test_Time_LLMs}.}
\end{abstract}


%\ccsdesc[300]{Do Not Use This Code~Generate the Correct Terms for Your Paper}
%\ccsdesc{Do Not Use This Code~Generate the Correct Terms for Your Paper}
%\ccsdesc[100]{Do Not Use This Code~Generate the Correct Terms for Your Paper}


\received{27 June 2025}

\maketitle

\input{draft/1-introduction}

\input{draft/2-background}

\input{draft/3-TTA}

\input{draft/4-TTR}
\input{draft/5-advanced}
\input{draft/5-future}
\input{draft/7-benchmark}
\input{draft/6-conclusion}

%\begin{acks}
%To Robert, for the bagels and explaining CMYK and color spaces.
%\end{acks}

\bibliographystyle{ACM-Reference-Format}
\bibliography{sample-base}

\appendix

\end{document}

%% file: draft/1-introduction.tex
\section{Introduction}

%Over the past decades, advances in deep learning have driven an AI revolution, enabling many tasks once considered unattainable to achieve significant breakthroughs in practical applications such as image recognition, sentiment analysis, machine translation, etc~\citep{he2016deep,NIPS2017_3f5ee243}. 
%The success of deep learning lies in the scaling effect, i.e. regardless of neural network architecture or data modality, larger models and more training data consistently lead to better performance on downstream tasks~\citep{hestness2017deeplearningscalingpredictable}.
%Particularly in the text modality, large language models (LLMs) represented by the GPT series~\citep{radfordimproving,radford2019language,brown2020language,ouyang2022traininglanguagemodelsfollow,OpenAI2023GPT4TR} have demonstrated the immense power of scaling effects.
Over the past decades, deep learning with its scaling effects has been the driving engine behind the artificial intelligence revolution.
Particularly in the text modality, large language models (LLMs) represented by the GPT series~\citep{brown2020language,ouyang2022traininglanguagemodelsfollow,OpenAI2023GPT4TR} have demonstrated that larger models and more training data lead to better performance on downstream tasks.
However, on the one hand, further scaling in the training phase becomes difficult due to the scarcity of data and computational resources~\citep{villalobos2024rundatalimitsllm}; on the other hand, existing models still perform far below expectations in terms of robustness and handling complex tasks.
These shortcomings are attributed to the model's reliance on fast, intuitive System-1 thinking, rather than slow, deep System-2 thinking~\citep{weston20232attentionisneed}.
Recently, large reasoning models (LRMs), represented by OpenAI-o1/o3~\citep{o12024}, DeepSeek-R1~\citep{guo2025deepseek}, and Gemini 2.5~\citep{google2025gemini25}, equipped with System-2 thinking, have gained attention for their outstanding performance in complex reasoning tasks. 
It demonstrates a test-time compute scaling effect: the greater the computational effort in the inference, the better the model's performance.

\begin{figure*}
    \centering
    \includegraphics[width=0.95\linewidth]{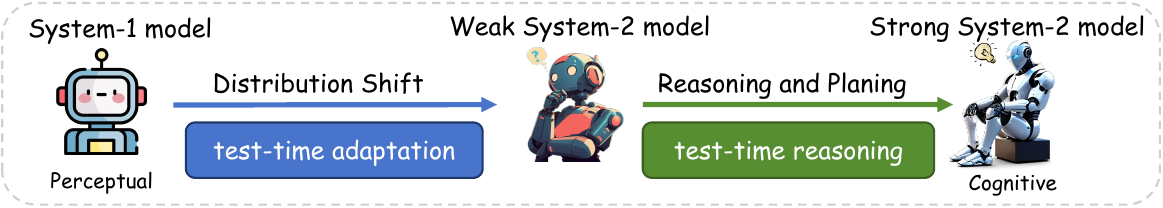}
    \caption{Illustration of test-time compute in the System-1 and System-2 model.}
    \label{fig:framework}
\end{figure*}

The concept of test-time compute emerged before the rise of LLMs and was initially applied to System-1 models (illustrated in Figure \ref{fig:framework}). % to enhance their robustness.
These System-1 models can only perform limited perceptual tasks, relying on patterns learned during training for predictions. As a result, they are constrained by the assumption that training and testing are identically distributed and lack robustness and generalization to distribution shifts~\citep{zhuang2020comprehensive}.
Many works have explored test-time adaptation (TTA) to improve model robustness by updating parameters~\citep{wang2021tent,ye-etal-2023-multi}, modifying the input~\citep{dong-etal-2024-survey}, editing representations~\citep{rimsky-etal-2024-steering}, and calibrating the output~\citep{zhang2023adanpc}.
With TTA, the System-1 model slows down its thinking process and has better generalization.
However, TTA is an implicit slow thinking, unable to exhibit explicit, logical thinking process like humans, and struggles to handle complex reasoning tasks.
Thus, TTA-enabled models perform weak System-2 thinking.

Currently, advanced LLMs with chain-of-thought (CoT) prompting~\citep{wei2022chain} have enabled language models to perform explicit System-2 thinking~\citep{hagendorff2023human}.
%, demonstrating the human-like cognitive ability to decompose problems and reason step-by-step. 
%However, they still struggle with complex tasks like reasoning and planning~\citep{stechly2024chain,sprague2024cotcotchainofthoughthelps}.
However, vanilla CoT is limited by error accumulation and linear thinking pattern~\citep{stechly2024chain}, making it difficult to fully simulate non-linear human cognitive processes such as brainstorming, reflection, and backtracking.
To achieve stronger System-2 models, researchers employ test-time compute strategies to extend model reasoning's breadth, depth and accuracy, such as repeated sampling~\citep{cobbe2021trainingverifierssolvemath}, self-correction~\citep{shinn2023reflexionlanguageagentsverbal}, and tree search~\citep{NEURIPS2023_271db992}.
Repeated sampling simulates the diversity of human thinking, self-correction enables LLMs to reflect, and tree search enhances reasoning depth and backtracking.

To the best of our knowledge, this paper is the first to systematically review test-time compute methods and thoroughly explore their critical role in advancing models from System-1 to weak System-2, and ultimately to strong System-2 thinking.
%The remainder of this paper provides a comprehensive survey of the latest research developments of test-time computing in System-1 and System-2 models.
In Section \ref{background}, we present the background of System-1 and System-2 thinking.
Section \ref{tta} and Section \ref{ttr} detail the test-time compute methods for the System-1 and System-2 models. Then, we discuss advanced topics and future directions in Section \ref{advanced} and \ref{future}. Additionally, we review benchmarks in Section \ref{app-benchmark}.

%% file: draft/2-background.tex
\section{Background}
\label{background}
\input{table/mind}
%\subsection{System-1 and System-2 Thinking}
System-1 and System-2 thinking are psychological concepts~\citep{kahneman2011thinking}. When recognizing familiar patterns or handling simple problems, humans often respond intuitively. 
This automatic, fast thinking is called System-1 thinking. 
In contrast, when dealing with complex problems like mathematical proofs or logical reasoning, deep and deliberate thought is required, referred as System-2 thinking—slow and reflective.
In the field of artificial intelligence, researchers also use these terms to describe different types of models~\citep{lecun2022path}. 
System-1 models respond directly based on internally encoded perceptual information and world knowledge without showing any intermediate decision-making process. 
In contrast, System-2 models explicitly generate reasoning processes and solve tasks incrementally.
Before the rise of LLMs, System-1 models were the mainstream in AI. 
Although many deep learning models, such as ResNet, Transformer, and BERT, achieve excellent performance in various tasks in computer vision and natural language processing, these System-1 models, similar to human intuition, lack sufficient robustness and are prone to errors~\citep{geirhos2020shortcut,wang-etal-2022-measure,du2023shortcut}.
Nowadays, the strong generation and reasoning capabilities of LLMs make it possible to build System-2 models. \citet{wei2022chain} propose the CoT, which allows LLMs to generate intermediate reasoning steps progressively during inference. Empirical and theoretical results show that this approach significantly outperforms methods that generate answers directly~\citep{kojima2022large,zhou2023leasttomost,feng2024towards,li2024chain}.
However, current System-2 models represented by CoT prompting still have shortcomings. The intermediate processes generated by LLMs may contain errors, leading to cumulative mistakes and ultimately resulting in incorrect answers. %~\citep{ling2023deductive,zhou2023leasttomost}.
Although retrieval-augmented generation (RAG) helps mitigate factual errors~\citep{trivedi-etal-2023-interleaving,guan2024mitigating,wang2024ratretrievalaugmentedthoughts}, their impact on improving reasoning abilities remains limited. As a result, CoT-enabled LLMs are still at the weak system-2 thinking stage.

%% file: table/mind.tex
\tikzstyle{my-box}=[
    rectangle,
    draw=hidden-black,
    rounded corners,
    text opacity=1,
    minimum height=1.5em,
    minimum width=5em,
    inner sep=2pt,
    align=center,
    fill opacity=.5,
]
\tikzstyle{leaf}=[
    my-box, 
    minimum height=1.5em,
    % fill=hidden-orange!60, 
    fill=hidden-green!50, 
    text=black,
    align=left,
    font=\normalsize,
    inner xsep=2pt,
    inner ysep=4pt,
]
\begin{figure*}[t]
    \vspace{-2mm}
    \centering
    \resizebox{\textwidth}{!}{
        \begin{forest}
            forked edges,
            for tree={
                child anchor=west,
                parent anchor=east,
                grow'=east,
                anchor=west,
                base=left,
                font=\large,
                rectangle,
                draw=hidden-black,
                rounded corners,
                align=left,
                minimum width=4em,
                edge+={darkgray, line width=1pt},
                s sep=3pt,
                inner xsep=2pt,
                inner ysep=3pt,
                line width=0.8pt,
                ver/.style={rotate=90, child anchor=north, parent anchor=south, anchor=center},
            },
            where level=1{text width=7em,font=\normalsize,}{},
            where level=2{text width=7em,font=\normalsize,}{},
            where level=3{text width=10em,font=\normalsize,}{},
            where level=4{text width=12em,font=\normalsize,}{},
            [
                Test-time Computing, ver
                [
                    ~Test-time \\
                    ~Adaptation~(\S\ref{tta})
                    [
                        ~Parameter \\
                        ~Updating
                        [
                            Test-time Training
                            [   
                                ~TTT~\citep{sun2020test}{,} 
                                TTT++~\citep{liu2021ttt++}{,} CPT~\citep{zhu2024efficienttesttimeprompttuning}{; etc.}
                                , leaf, text width=34em
                            ]
                        ]
                        [
                            Fully TTA
                            [
                                ~Tent~\citep{wang2021tent}{,}
                                SAR~\citep{niu2023towards}{,} TPT~\citep{shu2022test}{,}
                                OIL~\citep{ye-etal-2022-robust}{,} Anti-CF~\citep{su-etal-2023-beware}{,} RLCF~\citep{zhao2024testtime}{; etc.}
                                , leaf, text width=34em
                            ]
                        ]
                    ]
                    [
                        ~Input  \\~Modification
                        [
                            Demonstration Selection
                            [
                            ~EPR~\citep{rubin-etal-2022-learning}{,}
                            UDR~\citep{li-etal-2023-unified}{,}
                              CQL~\citep{zhang-etal-2022-active}{,}
                             Entropy~\citep{lu-etal-2022-fantastically}{,}
                            MDL~\citep{wu-etal-2023-self}{,}  HiAR~\citep{wu2024exampleshighlevelautomatedreasoning}{; etc.}
                            , leaf, text width=34em
                            ]
                        ]
                        [
                            Demonstration Creation
                            [
                            ~SG-ICL~\citep{kim2022selfgeneratedincontextlearningleveraging}{,}
                            Self-ICL~\citep{chen-etal-2023-self}{,} DAIL~\citep{su-etal-2024-demonstration}{,}
                            Auto-CoT~\citep{zhang2023automatic}{,} DAWN-ICL~\citep{tang2024dawn}{; etc.}
                            , leaf, text width=34em
                            ]
                        ]
                    ]
                    [
                        ~Representation \\~Editing
                        [
                            ~ITI~\citep{li2023inferencetime}{,} ActAdd~\cite{turner2024steeringlanguagemodelsactivation}{,}  SEA~\citep{qiu2024spectral}{,} CAA~\citep{rimsky-etal-2024-steering}{; etc.}
                            , leaf, text width=45.6em 
                        ]
                    ]
                    [
                        ~Output \\~Calibration
                        [
                            ~$k$NN-MT~\citep{khandelwal2021nearest}{,} AdaNPC~\cite{zhang2023adanpc}{,} Bi-$k$NN~\citep{you-etal-2024-efficient}{; etc.}
                            , leaf, text width=45.6em 
                        ]
                    ]
                ]
                [
                    ~Test-time \\~Reasoning~(\S\ref{ttr})
                    [
                        ~Feedback \\~Modeling
                        [
                            Score-based
                            [   
                                ~\citet{bradley1952rank}{,} ORM~\citep{cobbe2021trainingverifierssolvemath}{,} 
                                PAV~\citep{setlur2024rewardingprogressscalingautomated}{,}
                                PRM~\citep{lightman2024lets}{,} OmegaPRM~\citep{luo2024improvemathematicalreasoninglanguage}{; etc.}
                                , leaf, text width=34em
                            ]
                        ]
                        [
                            Generative-based
                            [
                                ~LLM-as-a-Judge~\citep{zheng2023judging}{,}
                                Auto-J~\citep{li2024generative}{,}
                                Prometheus~\citep{kim2024prometheus,kim-etal-2024-prometheus}{,} Fennec~\citep{liang2024fennecfinegrainedlanguagemodel}{,}\\
                                ~GenRM~\citep{zhang2024generativeverifiersrewardmodeling}{,} CriticRM~\citep{yu2024selfgeneratedcritiquesboostreward}{; etc.}
                                , leaf, text width=34em
                            ]
                        ]
                    ]
                    [
                        ~Search \\~Strategies
                        [
                            Repeated Sampling
                            [   
                                ~SC-CoT~\citep{wang2023selfconsistency}{,} 
                                PROVE~\citep{toh2024votescountprogramsverifiers}{,}~\citet{cobbe2021trainingverifierssolvemath}{,}
                                DiVeRSe~\citep{li-etal-2023-making}{,} PRS~\citep{ye-ng-2024-preference}{; etc.}
                                , leaf, text width=34em
                            ]
                            [   
                                ~Improvement training: ReST~\citep{gulcehre2023reinforcedselftrainingrestlanguage}{,} 
                                vBoN~\citep{amini2024variationalbestofnalignment}{,}
                                ~BoNBoN\citep{gui2024bonbon}{,} ~\citet{chow2024inferenceawarefinetuningbestofnsampling}{; etc.}
                                , leaf, text width=34em
                            ]
                        ]
                        [
                            Self-correction
                            [
                                ~Self-debug~\citep{chen2024teaching}{,} RIC~\citep{kim2023languagemodelssolvecomputer}{,} 
                                Critic~\citep{gou2024critic}{,}
                                Shepherd~\citep{wang2023shepherdcriticlanguagemodel}{,} MAD~\citep{liang2024encouragingdivergentthinkinglarge}{,} IoE~\citep{li2024confidencemattersrevisitingintrinsic}{,} \\
                                ~Refiner~\citep{paul-etal-2024-refiner}{,} Reflexion~\citep{shinn2023reflexionlanguageagentsverbal}{,} \citet{du2023improvingfactualityreasoninglanguage}{,} ProgCo~\citep{song2025progcoprogramhelpsselfcorrection}{,} S1~\citep{muennighoff2025s1simpletesttimescaling}{; etc.}
                                , leaf, text width=34em
                            ]
                            [   
                                ~Improvement training: GLoRe~\citep{havrilla2024glorewhenwhereimprove}{,} 
                                SCoRe\citep{kumar2024traininglanguagemodelsselfcorrect}{,}
                                Self-correct~\citep{welleck2023generating}{,}\\ ~\citet{qu2024recursiveintrospectionteachinglanguage}{,}~\citet{zhang2024learnanswertraininglanguage}{,} GRPO~\citep{shao2024deepseekmathpushinglimitsmathematical}{,} T1~\citep{hou2025advancinglanguagemodelreasoning}{; etc.}
                                , leaf, text width=34em
                            ]
                        ]
                        [
                            Tree Search
                            [   
                                ~ToT~\citep{NEURIPS2023_271db992}{,} 
                                RAP~\citep{hao-etal-2023-reasoning}{,} rStar~\citep{qi2024mutualreasoningmakessmaller}{,} 
                                TS-LLM~\citep{feng2024alphazeroliketreesearchguidelarge}{,} AlphaMATH~\citep{chen2024alphamathzeroprocesssupervision}{; etc.}
                                , leaf, text width=34em
                            ]
                            [   
                                ~Improvement training: ReST-MCTS*~\citep{zhang2024restmctsllmselftrainingprocess}{,} 
                                \citet{qin2024o1replicationjourneystrategic}{,}\\
                                ~MCTS-DPO~\citep{xie2024montecarlotreesearch}{,} \citet{zhao2024marcoo1openreasoningmodels}{,}~\citet{zhang2024o1codero1replicationcoding}{; etc.}
                                , leaf, text width=34em
                            ]
                        ]
                    ]
                ]
                [
                    ~Future \\~Directions~(\S\ref{future})
                    [
                        ~Generalization
                        [
                            ~\citet{jia2024generalizing}{,} GRM~\citep{yang2024regularizing}{,} DogeRM~\citep{lin-etal-2024-dogerm}{,} GenPRM~\citep{zhao2025genprmscalingtesttimecompute}{,} RM-R1~\citep{chen2025rmr1rewardmodelingreasoning}{,} Search-o1~\citep{li2025searcho1agenticsearchenhancedlarge}{,} Deep Research~\citep{li2025webthinkerempoweringlargereasoning}{; etc.}
                            , leaf, text width=45.6em
                        ]
                    ]
                    [
                        ~Multi-modal
                        [
                            ~MM-CoT~\citep{zhang2024multimodal}{,} VoT~\citep{wu2024minds}{,} \citet{lee-etal-2024-multimodal}{,} LLaVA-CoT~\citep{xu2024llavacotletvisionlanguage}{,} VisualPRM~\citep{wang2025visualprmeffectiveprocessreward}{,} LMM-R1~\citep{peng2025lmmr1empowering3blmms}{; etc.}
                            , leaf, text width=45.6em
                        ]
                    ]
                    [
                        ~Efficient
                        [
                            ~\citet{damani2024learninghardthinkinputadaptive}{,} OSCA~\citep{zhang2024scalingllminferenceoptimized}{,} \citet{wang-etal-2024-reasoning-token}{,} CCoT~\citep{cheng2024compressedchainthoughtefficient}{,} DEER~\citep{yang2025dynamicearlyexitreasoning}{,} L1~\citep{aggarwal2025l1controllinglongreasoning}{,} O1-Pruner~\citep{luo2025o1prunerlengthharmonizingfinetuningo1like}{; etc.}
                            , leaf, text width=45.6em
                        ]
                    ]  
                    [
                        ~Scaling Law
                        [
                            ~\citet{brown2024largelanguagemonkeysscaling}{,} \citet{snell2024scalingllmtesttimecompute}{,} 
                            \citet{wu2024inferencescalinglawsempirical}{,} \citet{chen2024simpleprovablescalinglaw}{; etc.}
                            , leaf, text width=45.6em
                        ]
                    ]
                    [
                        ~Combination
                        [
                            ~Marco-o1~\citep{zhao2024marcoo1openreasoningmodels}{,} TTT~\citep{akyürek2024surprisingeffectivenesstesttimetraining}{,} 
                            HiAR-ICL~\citep{wu2024exampleshighlevelautomatedreasoning}{; etc.}
                            , leaf, text width=45.6em
                        ]
                    ]
                ]
            ]
        \end{forest}
    }
    %\vspace{-4mm}
    \caption{Taxonomy of test-time computing methods and future directions.}
    \label{fig:taxonomy}
    % \vspace{-3mm}
\end{figure*}

%% file: draft/3-TTA.tex
\section{Test-time Adaptation for System-1 Thinking}
\label{tta}
\begin{figure*}[t]
    \centering
    \includegraphics[width=0.95\linewidth]{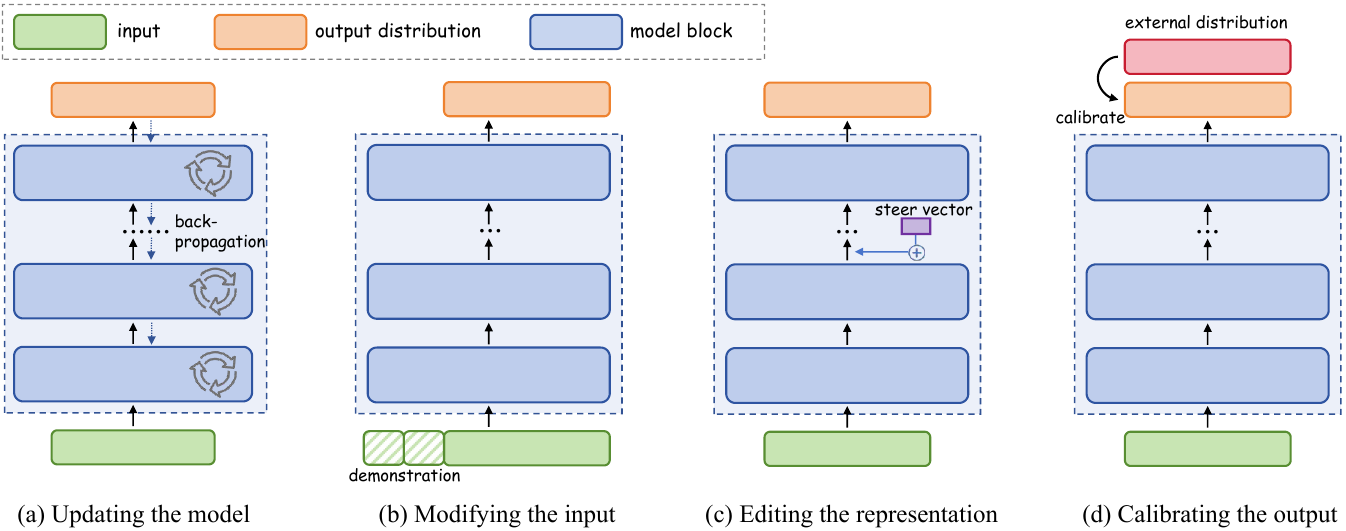}
    \caption{Illustration of various kinds of test-time adaptation methods.}
    \label{fig:tta}
\end{figure*}
\subsection{Updating the Model}
Model updating utilizes test sample information to finetune model parameters during the inference stage, enabling the model to adapt to the test distribution.
The key lies in obtaining test distribution information to provide learning signals and using appropriate parameters and optimization algorithms to achieve efficient and stable updates.
\paragraph{\textbf{Learning signal}}
%\subsubsection{Learning signal}
In the inference stage, the ground-truth of test samples is unavailable. Thus, many works attempt to design unsupervised or self-supervised objectives as learning signals.
Existing learning signals can be classified into two categories based on whether the training process can be modified: \textit{test-time training} (TTT) and \textit{fully test-time adaptation} (FTTA).
TTT assumes users can modify the training process by incorporating distribution-shift-aware auxiliary tasks. During test-time adaptation, the auxiliary task loss serves as the learning signal for optimization.
Many self-supervised tasks have been shown to be effective as auxiliary tasks in image modality, such as rotation prediction~\citep{sun2020test}, meta learning~\citep{bartler2022mt3}, masked autoencoding~\citep{gandelsman2022testtime} and contrastive learning
~\citep{liu2021ttt++,chen2022contrastive}.
Among them, contrast learning has been successfully applied to test-time adaptation for visual-language tasks due to its generalization of self-supervised learning within and across modalities~\citep{zhu2024efficienttesttimeprompttuning}.

In contrast, FTTA is free from accessing the training process and instead uses internal or external feedback on test samples as learning signals. 
Uncertainty is the most commonly learned signal, driven by the motivation that when test samples shift from the training distribution, the model's confidence in its predictions is lower, resulting in higher uncertainty.
Tent~\citep{wang2021tent} uses the entropy of model predictions as a measure of uncertainty and updates the model by minimizing the entropy.
MEMO~\citep{zhang2022memo} augments the data for a single test sample and then minimizes its marginal entropy, which is more stable compared to Tent in the single-sample TTA setting.
However, minimizing entropy also has pitfalls, as blindly reducing prediction uncertainty may cause the model to collapse and make trivial predictions~\citep{press2024entropyenigmasuccessfailure,zhao2023pitfalls,su-etal-2023-beware}.
Some works propose new regularization terms for minimizing entropy to avoid model collapse, including Kullback-Leibler divergence~\citep{su-etal-2023-beware}, moment matching~\citep{hassan2023align} and entropy matching~\citep{bar2024protectedtesttimeadaptationonline}.
For specific tasks, a small amount of human feedback or external model rewards can also serve as high-quality learning signals.
\citet{gao-etal-2022-simulating} and \citet{li-etal-2022-using} utilize user feedback to adapt the QA model. 
\citet{zhan-etal-2023-test} apply test-time adaptation to multilingual machine translation tasks by using COMET~\citep{rei-etal-2020-comet} for evaluating translation quality.
In cross-modal tasks such as image-text retrieval and image captioning, RLCF~\citep{zhao2024testtime} demonstrates its effectiveness by using CLIP scores~\citep{pmlr-v139-radford21a} as TTA signals.
In language modeling, training with relevant contextual text at test time can reduce perplexity~\citep{hardt2024testtime,wang2024with}. \citet{hubotter2025efficiently} theoretically shows that it reduces the uncertainty of test samples and proposes a more effective active learning selection strategy.

\paragraph{\textbf{Updating parameters}}
To advance the application of TTA in real-world scenarios, researchers must address challenges of efficiency and stability. To improve efficiency, many methods only fine-tune a small subset of parameters, such as normalization layers~\citep{schneider2020improving}, soft prompt~\citep{lester-etal-2021-power,shu2022test,hassan2023align,ma2023swapprompt,feng2023diverse,pmlr-v235-niu24a}, low-rank module~\citep{hu2022lora,imam2024testtimelowrankadaptation,hu2025testtimelearninglargelanguage}, adapter module~\citep{houlsby2019parameter,muhtar2024streamadapterefficienttesttime,su-etal-2023-beware} and cross-modality projector~\citep{zhao2024testtime}.
Although the number of parameters to fine-tune is reduced, TTA still requires an additional backward propagation. 
Typically, the time cost of a backward propagation is approximately twice that of a forward propagation.
Thus, \citet{pmlr-v235-niu24a} propose FOA, which is free from backward propagation by adapting soft prompt through covariance matrix adaptation evolution strategy.

The stability of TTA is primarily shown in two aspects.
On the one hand, unsupervised or self-supervised learning signals inevitably introduce noise into the optimization process, resulting in TTA optimizing the model in the incorrect gradient direction.
To address this, \citet{niu2023towards} and \citet{gong2024sotta} propose noise data filtering strategies and the robust sharpness-aware optimizer. 
On the other hand, in real-world scenarios, the distribution of test samples may continually shift, but continual TTA optimization may lead to catastrophic forgetting of the model’s original knowledge.
Episodic TTA~\citep{shu2022test,zhao2024testtime} is a setting to avoid forgetting, which resets the model parameters to their original state after TTA on a single test sample. 
However, episodic TTA frequently loads the original model, leading to higher inference latency and also limiting the model's incremental learning capability.
To overcome the dilemma, a common trick is the exponential moving average~\citep{wortsman2022model,ye-etal-2022-robust}, which incorporates information from previous model states.

\subsection{Modifying the Input}
When it comes to LLM, the large number of parameters makes model update-based TTA methods face a tougher dilemma of efficiency and stability.
As a result, input-modification-based methods, which do not rely on parameter updates, have become the mainstream method for TTA in LLMs.
The effectiveness of input-modified TTA stems from the in-context learning (ICL) capability of LLM, which can significantly improve the performance by adding some demonstrations before the test sample.
ICL is highly sensitive to the selection and order of demonstrations. Therefore, the core objective of input-modification TTA is to select appropriate demonstrations for the test samples and arrange them in the optimal order to maximize the effectiveness of ICL.
%ICL conducts implicit gradient descent

First, empirical studies~\citep{liu-etal-2022-makes} show that the more similar the demonstrations are to the test sample, the better the ICL performance. 
Therefore, retrieval models like BM25 and SentenceBERT are used to retrieve demonstrations semantically closest to the test sample and rank them in descending order of similarity~\citep{qin-etal-2024-context,luo2023dricldemonstrationretrievedincontextlearning}.
To improve the accuracy of demonstration retrieval, \citet{rubin-etal-2022-learning} and \citet{li-etal-2023-unified} specifically train the demonstration retriever by contrastive learning.
Then, as researchers delve deeper into the mechanisms of ICL, ICL is considered to conduct implicit gradient descent on the demonstrations~\citep{dai-etal-2023-gpt}.
Therefore, from the perspective of training data, demonstrations also need to be informative and diverse~\citep{su2022selectiveannotationmakeslanguage,li-qiu-2023-finding}.
\citet{wang2023large} view language models as topic models and formulate the demonstration selection problem as solving a Bayesian optimal classifier.
Additionally, the ordering of examples is another important area for improvement.
\citet{lu-etal-2022-fantastically} and \citet{wu-etal-2023-self} use information theory as a guide to select the examples with maximum local entropy and minimum description length for ranking, respectively. 
%Experimental results show that the minimum description length principle effectively guides example selection and ordering.
\citet{scarlatos2024reticlsequentialretrievalincontext} and \citet{zhang-etal-2022-active} consider the sequential dependency among demonstrations, and model it as a sequential decision problem and optimize demonstration selection and ordering through reinforcement learning.

Another line of work~\citep{chen-etal-2023-self,lyu-etal-2023-z,kim2022selfgeneratedincontextlearningleveraging,zhang2023automatic} argues that in practice, combining a limited set of externally provided examples may not always be the optimal choice. LLMs can leverage their generative and annotation capabilities to create better demonstrations.
DAIL~\citep{su-etal-2024-demonstration} constructs a demonstration memory, storing previous test samples and their predictions as candidate demonstrations for subsequent samples.
DAWN-ICL~\citep{tang2024dawn} further models the traversal order of test samples as a planning task and optimizes it by the Monte Carlo tree search (MCTS).

\subsection{Editing the Representation}
For generative LLMs, some works have found that the performance bottleneck is not in encoding world knowledge, but in the large gap between the information in intermediate layers and the output. During the inference phase, editing the representation can help externalize the intermediate knowledge into the output.
%The general form of representation editing is as follows:
%\begin{equation}
%    \hat{x}^{l}=x^l+\alpha r^l,
%\end{equation}
%where $x^l$ is the $l$-th layer's activation, $r^l$ is the steering vector and $\alpha$ is the steering intensity.
PPLM~\citep{Dathathri2020Plug} performs gradient-based representation editing under the guidance of a small language model to control the style of outputs.
ActAdd~\citep{turner2024steeringlanguagemodelsactivation} selects two semantically contrastive prompts and calculates the difference between their representations as a steering vector, which is then added to the residual stream.
Representation editing based on contrastive prompts has demonstrated its effectiveness in broader scenarios, including instruction following~\citep{stolfo2024improvinginstructionfollowinglanguagemodels}, alleviating hallucinations~\citep{li2023inferencetime,arditi2024refusallanguagemodelsmediated}, reducing toxicity~\citep{liu2024incontextvectorsmakingcontext,lu2024investigatingbiasrepresentationsllama} and personality~\citep{cao2024personalizedsteeringlargelanguage}.
SEA~\citep{qiu2024spectral} projects representations onto directions with maximum covariance with positive prompts and minimum covariance with negative prompts. They also introduce nonlinear feature transformations, allowing representation editing to go beyond linearly separable representations.
\citet{scalena-etal-2024-multi} conduct an in-depth study on the selection of steering intensity. They find that applying a gradually decreasing steering intensity to each output token can improve control over the generation without compromising quality.

\subsection{Calibrating the Output}
Using external information to calibrate the model's output distribution is also an efficient yet effective test-time adaptation method~\citep{Khandelwal2020Generalization}.
AdaNPC~\citep{zhang2023adanpc} designs a memory pool to store training data. During inference, given a test sample, AdaNPC recalls $k$ samples from the memory pool and uses a $k$NN classifier to predict the test sample. It then stores the test sample and its predicted label in the memory pool. Over time, the sample distribution in the memory pool gradually aligns with the test distribution.
In NLP, the most representative application of such methods is $k$NN machine translation ($k$NN-MT).
$k$NN-MT~\citep{khandelwal2021nearest} constructs a datastore to store contextual representations and their corresponding target tokens. During translation inference, it retrieves the $k$-nearest candidate tokens from the datastore based on the decoded context and processes them into probabilities. Finally, it calibrates the translation model's probability distribution by performing a weighted fusion of the model's probabilities and the retrieved probabilities.
$k$NN-MT has demonstrated superior transferability and generalization compared to traditional models in cross-domain and multilingual MT tasks. Subsequent studies have focused on improving its performance and efficiency~\citep{wang-etal-2022-efficient,zhu-etal-2023-knowledge,you-etal-2024-efficient} or applying its methods to other NLP tasks~\citep{wang2022knnnernamedentityrecognition,bhardwaj-etal-2023-knn}.

\paragraph{\textit{\textbf{Summary 1:}}}
\textit{Parameter updating and output calibration are the most versatile TTA methods. However, parameter updating suffers from training instability and inefficiency in LLMs, while output calibration relies on target domain information and risks knowledge leakage. Input modification and representation editing are free from training but have limited applicability: input modification is related to ICL capabilities, and representation editing demands manual prior knowledge.}

%% file: draft/4-TTR.tex
\section{Test-time Reasoning for System-2 Thinking}
\label{ttr}
\begin{figure*}[t]
    \centering
    \includegraphics[width=1\linewidth]{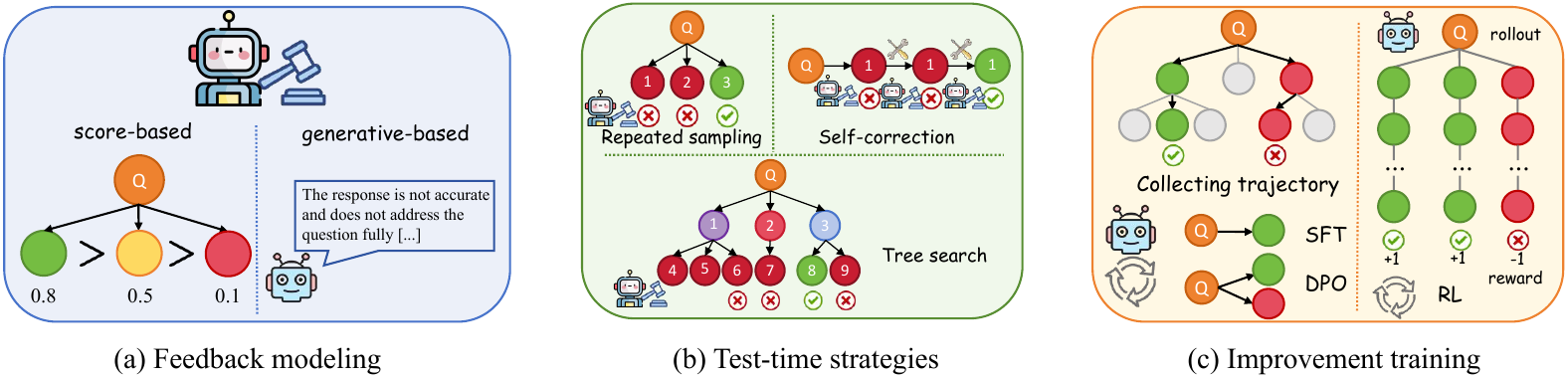}
    \caption{Illustration of feedback modeling, search strategies and improvement training in test-time reasoning.}
    \label{fig:ttr}
\end{figure*}
Test-time reasoning aims to spend more inference time to search for the most human-like reasoning process within the vast decoding search space.
In this section, we introduce the two core components of test-time reasoning: feedback modeling and search strategies (as shown in Figure \ref{fig:ttr}).
%, and then discuss how to improve test-time reasoning ability through training.
\subsection{Feedback Modeling}
\paragraph{\textbf{Score-based Feedback}}

Score-based feedback, also known as the verifier, aims to predict scalar scores to evaluate the alignment of generated results with ground truth or human cognitive processes. 
Its training process is typically similar to the reward model in RLHF, using various forms of feedback signals and modeling it as a classification~\citep{cobbe2021trainingverifierssolvemath} or rank task~\citep{bradley1952rank,yuan2024advancingllmreasoninggeneralists,hosseini2024vstar}.
In reasoning tasks, verifiers are primarily categorized into two types: outcome-based verifiers (ORMs) and process-based verifiers (PRMs).
ORMs~\citep{cobbe2021trainingverifierssolvemath} use the correctness of the final CoT result as training signals.
\citet{liu2025acemathadvancingfrontiermath} provide a detailed recipe for training a strong ORM.
In contrast, PRMs~\citep{uesato2022solvingmathwordproblems,lightman2024lets,zhang2024entropy} are trained based on the correctness of each reasoning step.
Compared to ORMs, PRMs cannot only evaluate intermediate reasoning steps but also assess the entire reasoning process more precisely.
%PRMs not only evaluate intermediate reasoning steps but also evaluate the entire reasoning process more accurately than ORMs. 
However, PRMs require more human effort to annotate feedback for the intermediate steps.
Math-Shepherd~\citep{wang-etal-2024-math}, OmegaPRM~\citep{luo2024improvemathematicalreasoninglanguage} and EpicPRM~\citep{sun2025efficientprecisetrainingdata} utilize MCTS algorithm to collect high-quality process supervision data automatically.
\citet{zhang2025lessons} utilize critic models to evaluate process annotations collected by MCTS, filtering out low-quality data to improve the training effectiveness of PRMs.
\citet{setlur2024rewardingprogressscalingautomated} argue that PRMs should evaluate the advantage of each step for subsequent reasoning rather than focusing solely on its correctness. 
They propose process advantage verifiers (PAVs) and efficiently construct training data through MCTS.
Furthermore, \citet{lu2024autopsv} and \citet{yuan2024free} notice that ORMs implicitly model the advantage of each step, leading them to automatically annotate process supervision data using ORMs or directly train PRMs on outcome labels, respectively.
\citet{liu2025adaptivestepautomaticallydividingreasoning} adaptively segment reasoning steps based on tokens with higher uncertainty, which is used to train better PRMs.

% Score-based feedback modeling overlooks the generative capabilities of LLMs, making it difficult to detect fine-grained errors. Thus, recent works propose generative score-based verifiers~\citep{ankner2024critiqueoutloudrewardmodels,ye2024improvingrewardmodelssynthetic,chiang2025tractregressionawarefinetuningmeets,she2025rprmreasoningdrivenprocessreward}.
% GenRM~\citep{zhang2024generativeverifiersrewardmodeling} leverages instruction tuning to enable the verifier to answer `Is the answer correct (Yes/No)?' and uses the probability of generated `Yes' token as the score. GenRM can also incorporate CoT, allowing the verifier to generate the corresponding rationale before answering `Yes' or `No'.
% ThinkPRM~\citep{khalifa2025processrewardmodelsthink} evaluates each reasoning step with long CoT and requires only 1\% of the process supervision data compared to discriminative PRMs.
% Critic-RM~\citep{yu2024selfgeneratedcritiquesboostreward} jointly trains the critique model and the verifier.
% During inference, the verifier scores according to answers and verbal-based feedback generated by the critique model.

\paragraph{\textbf{Generative-based Feedback}}
Although the verifier can evaluate the correctness of generated answers or steps, it lacks interpretability, making it unable to locate the specific cause of errors or provide correction suggestions.
Generative-based feedback, also referred to critic, fully leverages the LLM's generative and instruction-following ability~\citep{ankner2024critiqueoutloudrewardmodels,ye2024improvingrewardmodelssynthetic,chiang2025tractregressionawarefinetuningmeets,she2025rprmreasoningdrivenprocessreward}. By designing specific instructions, it can perform pointwise or pairwise evaluation from multiple dimensions, and even provide suggestions for revision in natural language.
Powerful closed-source LLMs, such as GPT-4 and Claude, are effective critics. They 
can perform detailed and controlled assessments of generated texts, such as factuality, logical errors, coherence, and alignment, with high consistency with human evaluations~\citep{wang-etal-2023-chatgpt,luo2023chatgptfactualinconsistencyevaluator,liu-etal-2023-g,chiang-lee-2023-large}.
However, they still face biases such as length, position, and perplexity~\citep{bavaresco2024llmsinsteadhumanjudges,wang-etal-2024-large-language-models-fair,stureborg2024largelanguagemodelsinconsistent}.
LLM-as-a-Judge~\citep{zheng2023judging} carefully designs system instructions to mitigate the interference of biases.
BSM~\citep{saha-etal-2024-branch} evaluates based on multiple criteria and then merges them.
\citet{peng2025agenticrewardmodelingintegrating} employ multi-agents to jointly evaluate answers' factuality and instruction-following.

To obtain cheaper verbal-based feedback, open-source LLMs can also serve as competitive alternatives through supervised fine-tuning ~(SFT)~\citep{wang2024pandalm,zhu2023judgelmfinetunedlargelanguage,liang2024fennecfinegrainedlanguagemodel,paul-etal-2024-refiner}.
Shepherd~\citep{wang2023shepherdcriticlanguagemodel} collects high-quality training data from human annotation and online communities to fine-tune an evaluation model.
Auto-J~\citep{li2024generative} collects queries and responses from various scenarios and designs evaluation criteria for each scenario. GPT-4 then generates critiques of the responses based on these criteria and distills its critique ability to open-source LLMs.
Prometheus~\citep{kim2024prometheus,kim-etal-2024-prometheus} designs more fine-grained evaluation dimensions. It trains a single evaluation model and a pairwise ranking model separately, then unifies them into one LLM by weight merging.
%Fennec~\citep{liang2024fennecfinegrainedlanguagemodel} allows GPT-4 to determine the evaluation criteria for each query, and generate corresponding verbal feedback. 
%Compared to previous work, Fennec's evaluation criteria are more flexible, the generated evaluation data is more diverse, and it aligns better with human behavior.
To reduce reliance on human annotations and external LLMs, \citet{wang2024selftaughtevaluators} propose a self-training method: the critique model generates positive and negative responses, then collects critique data via rejection sampling to perform iterative finetuning.
Building on self-training, EvalPlanner enables~\citep{saha2025learningplanreason} the critique model to plan evaluation processes and criteria, conduct critiques based on these, and then collect positive and negative samples to improve the critique model via DPO~\citep{rafailov2023direct}.
\citep{yu2025improvellmasajudgeabilitygeneral} carefully synthesize and filter data, enabling the base model to achieve strong critique abilities with only 40K samples for SFT and DPO training.
GenRM~\citep{zhang2024generativeverifiersrewardmodeling} leverages instruction tuning to enable the verifier to answer `Is the answer correct (Yes/No)?' and uses the probability of generated `Yes' token as the score. GenRM can also incorporate CoT, allowing the verifier to generate the corresponding rationale before answering `Yes' or `No'.
ThinkPRM~\citep{khalifa2025processrewardmodelsthink} evaluates each reasoning step with long CoT and requires only 1\% of the process supervision data compared to discriminative PRMs.
Critic-RM~\citep{yu2024selfgeneratedcritiquesboostreward} jointly trains the critic and the verifier.
%During inference, the verifier scores according to answers and verbal-based feedback generated by the critique model.
\input{table/reward}

\input{table/catagory}
\subsection{Search Strategies}
\subsubsection{\textbf{Repeated Sampling}}
\ 
\newline
Sampling strategies such as top-p and top-k are commonly used decoding algorithms in LLM inference. They introduce randomness during decoding to enhance text diversity, allowing for parallelly sampling multiple generated texts.
Through repeated sampling, we have more opportunities to find the correct answer.
Repeated sampling is particularly suitable for tasks that can be automatically verified, such as code generation, where we can easily identify the correct solution from multiple samples using unit tests~\citep{li2022competition,rozière2024codellamaopenfoundation}.
For tasks that are difficult to verify, like math word problems, the key to the effectiveness of repeated sampling is the verification strategy.

\paragraph{\textbf{Verification strategy}}
Verification strategies include two types: majority voting and best-of-N (BoN) sampling.
\textit{Majority voting}~\citep{li2024agentsneed,lin-etal-2024-just} selects the most frequently occurring answer in the samples as the final answer, which is motivated by ensemble learning.
Majority voting is simple yet effective. For instance, self-consistency CoT~\citep{wang2023selfconsistency} can improve accuracy by 18\% over vanilla CoT in math reasoning tasks.
However, the majority does not always hold the truth, as they may make similar mistakes. 
Therefore, some studies perform filtering or weighting before voting. For example, the PROVE framework~\citep{toh2024votescountprogramsverifiers} converts CoT into executable programs, filtering out samples if the program's results are inconsistent with the reasoning chain's outcomes.
\citet{kang2025scalablebestofnselectionlarge} propose the self-certainty metric to weight votes based on their ranking.
\citet{huang2025efficienttesttimescalingselfcalibration} use the calibrated confidence as vote weights, and adjust the number of samples based on response agreement to improve efficiency~\citep{aggarwal-etal-2023-lets}.
RPC~\citep{zhou2025bridginginternalprobabilityselfconsistency} demonstrates theoretically and experimentally that filtering low-probability samples improves the performance and efficiency of majority voting.

\textit{Best-of-N sampling} uses a verifier to score each response and selects the one with the highest score as the final answer~\citep{stiennon2020learning,cobbe2021trainingverifierssolvemath,nakano2022webgptbrowserassistedquestionansweringhuman,liu2025bagtricksinferencetimecomputation}.
\citet{li-etal-2023-making} propose a voting-based BoN variant, which performs weighted voting on all answers based on the verifier's scores and selects the answer with the highest score.
\citep{liu2025pairwise} design BoN in a knockout tournament, using pairwise comparison verifiers to filter out the best response.
In addition, some works aim to improve the efficiency of BoN.
Inspired by speculative decoding, \citet{zhang2024accelerating,qiu2024treebonenhancinginferencetimealignment,sun2024fast,yu2025scalingflawsverifierguidedsearch} and \citet{manvi2024adaptiveinferencetimecomputellms} evaluate each reasoning step and prune low-scoring sampled results, halting further generation for those paths, thereby significantly reducing the overall time cost.
PRS~\citep{ye-ng-2024-preference} enables LLMs to self-critique and self-correct, guiding the model to generate expected responses with fewer sampling times.
\citet{li2025rewardingcurseanalyzemitigate} compare BoN and majority voting, demonstrating that BoN is suitable for harder questions with moderate diversity of response distribution.

\paragraph{\textbf{Improvement Training}}
Repeated sampling has proven to be a simple yet effective method, even surpassing models fine-tuned with RLHF~\citep{gao2023scaling,hou2024doesrlhfscaleexploring}. However, it costs much inference time that is difficult to afford in practical applications.
Therefore, many studies have attempted to train the model by BoN sampling to approximate the BoN distribution, thereby reducing the search space during inference.
ReST~\citep{gulcehre2023reinforcedselftrainingrestlanguage} samples responses with reward values above a threshold from the policy model as self-training data and fine-tunes the policy model by offline reinforcement learning. In each iteration, ReST samples new training data.
vBoN~\citep{amini2024variationalbestofnalignment}, BoNBoN~\citep{gui2024bonbon} and BOND~\citep{sessa2024bondaligningllmsbestofn} derive the BoN distribution and minimize the difference between the policy model's distribution and the BoN distribution.
\citet{chow2024inferenceawarefinetuningbestofnsampling} design a BoN-aware loss to make the policy model more exploratory during fine-tuning.

%\textcolor{red}{rl}: \citep{kazemnejad2024vineppo}

\subsubsection{\textbf{Self-correction}}
\ 
\newline
Self-correction is a sequential test-time compute method that enables LLMs to iteratively revise and refine generated results based on external or internal feedback~\citep{shinn2023reflexionlanguageagentsverbal}.
\begin{figure*}[t]
    \centering
    \includegraphics[width=0.95\linewidth]{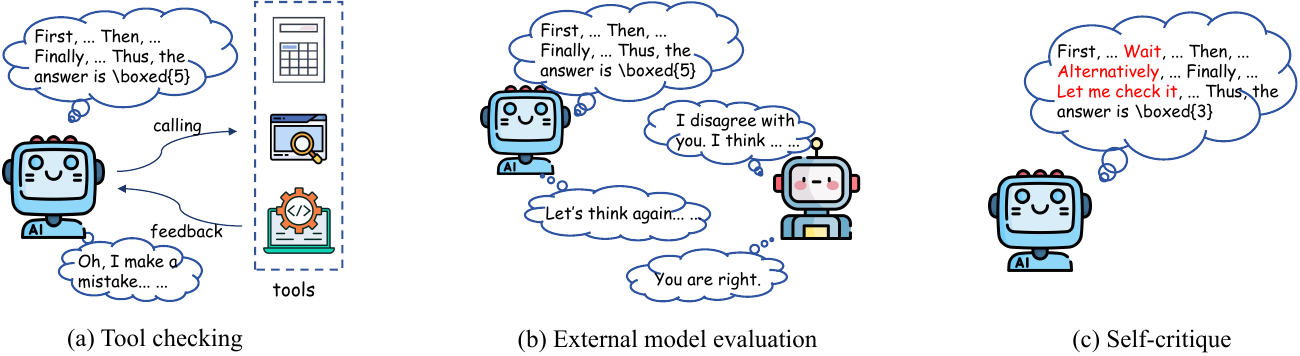}
    \caption{Illustration of self-correction with tool checking, external model evaluation, and self-critique.}
    \label{fig:correct}
\end{figure*}

\paragraph{\textbf{Feedback sources}}
The feedback used for self-correction is typically presented in natural language and comes from various sources, including human evaluation, tool checking, external model evaluation, and intrinsic feedback.
\textit{Human evaluation} is the gold standard for feedback, but due to its high cost and limited scalability, it is mainly used in early research to explore the upper limits of self-correction capabilities~\citep{tandon2021interscriptdatasetinteractivelearning,elgohary-etal-2021-nl,tandon-etal-2022-learning}.
For certain domain-specific tasks, \textit{external tool checking} provides accurate and efficient feedback~\citep{gou2024critic,chen2024teaching,gao-etal-2023-rarr}. For example, \citet{Yasunaga20DrRepair} propose to obtain feedback from compilers in code repair and generation tasks.
In embodied tasks, the environment can provide precise feedback on the action trajectories of LLM-based agents~\citep{wang-etal-2024-e2cl}.

\textit{External model evaluation} is an effective feedback source for general tasks, such as various verbal-based critique models described in Section 4.1.
For example, \citet{paul-etal-2024-refiner} first define multiple error types for natural language reasoning tasks and then design the corresponding feedback templates. They train an evaluation model using synthetic feedback training data, and with the critic, the reasoning model achieves substantial performance improvement.
Multi-agent debate~\citep{du2023improvingfactualityreasoninglanguage,xiong-etal-2023-examining,liang2024encouragingdivergentthinkinglarge,chen-etal-2024-reconcile,wang2024mixtureofagentsenhanceslargelanguage} is another mechanism that leverages external feedback to enhance reasoning capabilities. 
In this approach, models do not have distinct roles as reasoners and critics. Instead, multiple models independently conduct reasoning, critique each other, and defend or refine their reasoning based on feedback.
This process continues until agents reach a consensus or a judge model summarizes the final reasoning results.
The multi-agent debate has shown its potential in fact-checking~\citep{kim2024llmsproducefaithfulexplanations,khan2024debating}, commonsense QA~\citep{xiong-etal-2023-examining}, faithful evaluations~\citep{chan2024chateval}, and complex reasoning~\citep{du2023improvingfactualityreasoninglanguage}.
However, multi-agent debate may be unstable, as LLMs are susceptible to adversarial information and may revise correct answers to incorrect ones in response to misleading inputs~\citep{laban2024surechallengingllmsleads,amayuelas2024multiagentcollaborationattackinvestigating}. 
Therefore, a successful multi-agent debate requires that LLMs maintain their stance when faced with incorrect answers from other models while remaining open to valid suggestions~\citep{stengeleskin2024teachingmodelsbalanceresisting}.
In general, the more LLMs involved in the debate, the stronger the overall reasoning performance. However, this significantly increases the number of LLM inferences required, and the length of input context, posing a major challenge to LLM inference costs~\citep{liu2024groupdebateenhancingefficiencymultiagent}.
To reduce debate inference costs,
\citet{li2024improvingmultiagentdebatesparse} investigate the impact of topological connections among multiple agents and show that sparse connections, such as ring structures, are not inferior to the fully connected topology.
GroupDebate~\citep{liu2024groupdebateenhancingefficiencymultiagent} divides LLMs into groups that conduct debates internally and only share the consensus results between groups.

\textit{Self-critique} assumes that LLMs can self-evaluate their outputs and optimize them through intrinsic feedback~\citep{yuan2024selfrewardinglanguagemodels}.
This idea stems from a fundamental principle in computational complexity theory: verifying whether a solution is correct is typically easier than solving the problem.
\citet{bai2022constitutionalaiharmlessnessai} propose self-correcting harmful responses from LLMs by prompting themselves. Self-Refine~\citep{madaan2023selfrefine} and RCI Prompting~\citep{kim2023languagemodelssolvecomputer} iteratively prompt LLMs to self-correct their responses in tasks such as arithmetic reasoning.
IoE~\citep{li2024confidencemattersrevisitingintrinsic} observes that LLMs may over-criticize themselves during self-critique, leading to performance degradation, and designs prompt to guide LLMs in assessing confidence.
ProgCo~\citep{song2025progcoprogramhelpsselfcorrection} leverage the advantages of code in expressing complex logic, enabling LLMs to generate responses in pseudo-code form, followed by self-critique and refinement.
SETS~\citep{chen2025setsleveragingselfverificationselfcorrection} combines the strengths of repeated sampling and self-critique, applying self-critique and correction to each sampled reasoning path and choosing the final solution via majority voting.
S1~\citep{muennighoff2025s1simpletesttimescaling} adds the ``wait'' to the reasoning process, prompting the LLM to critique its reasoning.

\paragraph{\textbf{Arguments}}
The effectiveness of self-correction, especially the self-critique, has remained controversial. Several empirical studies on code generation~\citep{olausson2024is}, commonsense QA~\citep{huang2024large}, math problem-solving~\citep{wang2024reasoningtokeneconomiesbudgetaware}, planning~\citep{valmeekam2023largelanguagemodelsreally}, and graph coloring~\citep{stechly2023gpt4doesntknowits} confirm that self-correction is not a guaranteed solution for improving performance.
%\citet{stechly2023gpt4doesntknowits} conduct experiments on graph coloring, an NP-complete reasoning problem, and conclude that the effectiveness of self-correction depends on the correctness of the feedback. For sufficiently complex reasoning tasks, even GPT-4 lacks the ability to verify correctness, so self-critique feedback does not improve all reasoning tasks.
%\citet{huang2024large} also observe abnormal performance drops during self-correction in math word problem-solving and commonsense QA tasks. They attribute the previous performance gains to ground-truth leakage, as LLMs cannot correctly determine when to stop the self-correction loop.
\citet{kamoi2024llmsactuallycorrectmistakes} think the effectiveness of self-correction has been overestimated. Previous successes either rely on oracle answers or weak initial answers. Only tasks that can be broken down into easily verifiable sub-tasks can truly benefit from self-correction. They suggest fine-tuning specific evaluation models to achieve better self-correction.
\citet{zhang2024understandingdarkllmsintrinsic} try to interpret and alleviate the failure of self-critique via human-like cognitive bias.
\citet{tyen-etal-2024-llms} decouple the abilities of LLMs to identify and correct errors and create the corresponding evaluation datasets. The evaluation results show that LLMs do not lack the ability to correct errors during self-correction, and their main performance bottleneck lies in locating the errors.
\citet{yang2024confidencevscritiquedecomposition} decompose self-critique into confidence and critique capabilities. Empirical studies show that fine-tuning is necessary to enhance both capabilities simultaneously, while prompt engineering can only achieve a trade-off.

\paragraph{\textbf{Improvement Training}}
Most of the aforementioned self-correction methods demonstrate significant performance improvements on advanced closed-source large models or open-source LLMs with over 70B parameters. 
However, for medium-scale open-source models with weaker capabilities, we need to further fine-tune them to unlock their self-correction capabilities.
Supervised fine-tuning optimizes the model using high-quality multi-turn correction data, either manually annotated~\citep{saunders2022selfcritiquingmodelsassistinghuman}, self-rationalize~\citep{zelikman2022star,yuan2025agentrtraininglanguagemodel}, multi-agent debate~\citep{subramaniam2025multiagent} or sampled from stronger LLMs~\citep{an2023learning,paul-etal-2024-refiner,qu2024recursiveintrospectionteachinglanguage,gao2024embeddingselfcorrectioninherentability,zhang2024learnanswertraininglanguage,xi2024enhancingllmreasoningcritique}.
GLoRe~\citep{havrilla2024glorewhenwhereimprove} considers that LLMs need global or local refinement for different types of errors. To address this, they construct training sets for global and local refinement, train verifiers to identify global and local errors, and develop LLMs for refinement based on different global or local feedback signals.
\citet{xi2024enhancingllmreasoningcritique} design a scalable framework for synthesizing self-correction training data, enabling reasoning models to generate controlled errors and receive feedback from critics to self-correct.
%PTR\citep{du2024thinkthriceactprogressive}
Although SFT is effective, training data from offline-generated self-correction trajectories can only simulate limited correction patterns. This leads to the distribution mismatch with the actual self-correction behavior during model inference. 
Self-correct~\citep{welleck2023generating} adopts online imitation learning, re-sampling new self-correction trajectories for training after each training epoch.
To further expand the exploration space of LLMs, many studies adopt flexible RL algorithms to surpass the performance limits of SFT.
SCoRe~\citep{kumar2024traininglanguagemodelsselfcorrect} proposes using the multi-turn RL method to improve self-critique and self-correction capability.
T1~\citep{hou2025advancinglanguagemodelreasoning} employs self-correction training data for SFT cold-start, followed by RL training using the RLOO algorithm~\citep{ahmadian-etal-2024-back}. During the RL phase, high-temperature sampling and entropy rewards encourage the LLM to explore more diverse reasoning paths.
Deepseek-R1~\citep{guo2025deepseek} uses rule-based rewards and the GRPO algorithm~\citep{shao2024deepseekmathpushinglimitsmathematical} for RL training. It also demonstrates RL's immense potential, even without SFT cold-start, its exploration capabilities suffice to endow LLMs with strong reasoning abilities.

\subsubsection{\textbf{Tree Searching}}
\ 
\newline
Repeated sampling and self-correction scale test-time compute in parallel and sequentially, respectively.
Human thinking is a tree search that combines brainstorming in parallel with backtracking to find other paths to solutions when it encounters a dead end.
Search algorithms and value functions are two critical components in tree searching.

%\begin{figure}
%    \centering
%    \includegraphics[width=0.8\linewidth]{latex/figure/mcts.pdf}
%    \caption{Caption}
%    \label{fig:mcts}
%\end{figure}

\paragraph{\textbf{Search algorithm}}
In LLM reasoning, current search algorithms include uninformed search and heuristic search.
Uninformed search explores the search space according to a fixed rule.
For example, tree-of-thought (ToT)~\citep{NEURIPS2023_271db992} adopts the BFS or DFS to search, while \citet{xie2023selfevaluation} use beam search.
Uninformed search is usually less efficient for problems with large search spaces, so heuristic search strategies represented by A$^{*}$~\citep{meng2024llmalargelanguagemodel,wang2024qimprovingmultistepreasoning} and MCTS~\citep{hao-etal-2023-reasoning,bi2024forestofthoughtscalingtesttimecompute,park2024ensembling} are widely used in reasoning tasks.
MCTS, which eliminates the need for explicit heuristics, leverages stochastic simulations and adaptive tree expansion under uncertain environments, making it well-suited for large state spaces. It optimizes search results gradually through four steps: selection, expansion, simulation, and backpropagation, approaching the optimal solution. In contrast, A$^*$ uses a heuristic function-guided deterministic search to guarantee optimal paths, but its performance depends on the design of the heuristic function. As a result, MCTS has been successfully applied to tasks such as RAG~\citep{hu-etal-2024-serts,jiang2024ragstarenhancingdeliberativereasoning,li2024thinkciteimprovingattributedtext,feng2025airragactivatingintrinsicreasoning}, QA~\citep{luo2025kbqao1agenticknowledgebase,gan2025mastermultiagentllmspecialized}, hallucinations mitigation~\citep{cheng2025thinkmorehallucinateless}, text-to-SQL~\citep{yuan2025mctssqleffectiveframeworktexttosql}, etc.
Additionally, \citet{long2023largelanguagemodelguided} trains an LLM controller using reinforcement learning to guide the LLM reasoner's search path, and \citet{chari2025pheromonebasedlearningoptimalreasoning} utilizes ant colony evolutionary algorithm to guide tree search.

\paragraph{\textbf{Value function}}
The value function evaluates the value of each state and guides the tree to expand towards branches with higher values in heuristic tree search.
\citet{xu2023traingainunleashmathematical} train an energy function by noise-contrastive estimation as the value function.
RAP~\citep{hao-etal-2023-reasoning} designs a series of heuristic value functions, including the likelihood of the action, the confidence of the state, self-evaluation results, and task-specific reward, and combines them according to task requirements.
Reliable and generalized value functions facilitate the application of MCTS to more complex problems with deeper search spaces.
AlphaMath~\citep{chen2024alphamathzeroprocesssupervision} and TS-LLM~\citep{feng2024alphazeroliketreesearchguidelarge} replace the hand-crafted value function with a learned LLM value function, automatically generating reasoning process and step-level evaluation signals in MCTS.
VerifierQ~\citep{qi2024verifierqenhancingllmtest}  integrates implicit Q-learning and contrastive Q-learning to train the value function, effectively mitigating the overestimation issue at the step level.
Traditional MCTS methods expand only one trajectory, while rStar~\citep{qi2024mutualreasoningmakessmaller} argues that the current value function struggles to guide the selection of the optimal path accurately. Therefore, rStar retains multiple candidate paths and performs reasoning with another LLM, ultimately selecting the path where both LLMs' reasoning results are consistent.
\citet{gao2024interpretablecontrastivemontecarlo} propose SC-MCTS, which combines multiple reward models, including contrastive reward, likelihood, and self-evaluation as value functions.
MCTSr~\citep{zhang2024accessinggpt4levelmathematical} and SR-MCTS~\citep{zhang2024llamaberrypairwiseoptimizationo1like} take complete responses as nodes, expanding the search space through self-critique and correction. SR-MCTS utilizes pairwise preference rewards and global quantile score as the value function, offering a more robust value function estimation.
%TS-LLM~\citep{feng2024alphazeroliketreesearchguidelarge} replaces the hand-craft value function with a learned LLM value function, which can evaluate each step so that the backup process can happen in the intermediate step, without the need for complete generations.
 
\paragraph{\textbf{Improvement Training}}
Tree search can guide LLMs to generate long reasoning processes, and these data help train LLMs with stronger reasoning abilities~\citep{zhai2024enhancingdecisionmakingllmagents,xu2024sramctsselfdrivenreasoningaugmentation,guan2025rstarmathsmallllmsmaster}.
ReST-MCTS*~\citep{zhang2024restmctsllmselftrainingprocess} uses process rewards as a value function to guide MCTS, collecting high-quality reasoning trajectories and the value of each step to improve the policy model and reward model.
Due to the step-by-step exploration of tree search, it can obtain finer-grained step-level feedback signals. MCTS-DPO~\citep{xie2024montecarlotreesearch} collects step-level preference data through MCTS and uses DPO for preference learning.
AlphaLLM-CPL~\citep{wang2024selfimprovementllmsmctsleveraging} ranks trajectories based on preference reward gaps and policy prediction gaps, employing curriculum learning to efficiently utilize MCTS-collected trajectories.
Recently, many LMRs~\citep{qin2024o1replicationjourneystrategic,zhao2024marcoo1openreasoningmodels,zhang2024o1codero1replicationcoding} have also confirmed the necessity of using tree search to construct high-quality long reasoning chain data for training.

\paragraph{\textit{\textbf{Summary 2:}}}
\textit{Repeated sampling is easy to implement and improves answer diversity, making it suitable for open-ended or easily verifiable tasks, though computationally inefficient. Self-correction relies on precise, fine-grained feedback and works well for easily verifiable tasks, but may not perform well with poor feedback or weak reasoning capability. Tree search optimizes complex planning tasks globally but involves complex implementation.}

%% file: table/reward.tex
\begin{table*}[t]
\centering

%\vspace{\baselineskip}
%\footnotesize
\resizebox{\textwidth}{!}{
\begin{tabular}{lllllc}
\toprule
Category & Sub-category & Representative Methods & Domain & Objective & Description  \\

\midrule
\multirow{7}*{Score-based} & \multirow{2}*{ORM} & \citet{cobbe2021trainingverifierssolvemath} & Math & classification & ORM; human annotated data \\
& & Acemath~\citep{liu2025acemathadvancingfrontiermath} & Math &  list-wise Bradley-Terry & ORM; sampling training data from multiple LLMs\\
\cmidrule(r){2-6}
& \multirow{5}*{PRM} & \citet{lightman2024lets} & Math & classification & PRM; human annotated data \\
& & Math-shepherd~\citep{wang-etal-2024-math} & Math & classification & PRM; annotating processes via MC estimation\\
& & \citet{zhang2025lessons} & Math & classification/regression & PRM; annotating processes via MCTS and LLM-as-a-judge\\
& & Implicit PRM~\citep{yuan2024free} & Math & implicit reward modeling & PRM; training PRMs with outcome labels\\
& & ASPRM~\citep{liu2025adaptivestepautomaticallydividingreasoning} & Math, Code & classification &  PRM; adaptive segment step; annotating processes via MC estimation\\
\midrule
\multirow{9}*{Generative-based} & \multirow{2}*{Training-free} & LLM-as-a-Judge~\citep{zheng2023judging} & General & - & Designing system instructions to mitigate biases\\
& & BSM~\citep{saha-etal-2024-branch} & General & - & Dividing into multiple criteria and then merging\\
\cmidrule(r){2-6}
& \multirow{7}*{Training-based} & Shepherd~\citep{wang2023shepherdcriticlanguagemodel} & General & SFT & Collecting data from human annotation and the Internet\\
& & Prometheus~\citep{kim-etal-2024-prometheus} & General & SFT & Training single and pairwise models and then merging them\\
& & EvalPlanner~\citep{saha2025learningplanreason} & General & DPO & Planing evaluation processes and then evaluating \\
& &GenRM~\citep{zhang2024generativeverifiersrewardmodeling} & Math & SFT & PRM; synthesizing critique data from external LLMs\\
& & R-PRM~\citep{she2025rprmreasoningdrivenprocessreward} & Math & SFT \& DPO & PRM; synthesizing critique data from external LLMs  \\
& & Critic-RM~\citep{yu2024selfgeneratedcritiquesboostreward} & General & SFT \& Bradley-Terry & ORM; synthesizing and filtering critique data via self-critique\\
& & CLoud~\citep{ankner2024critiqueoutloudrewardmodels} & General & SFT \& Bradley-Terry & ORM; synthesizing data from external LLMs and self-critique  \\
%& & TRACT~\citep{chiang2025tractregressionawarefinetuningmeets} & General & \\
\bottomrule
\end{tabular}
}
\caption{Overview of feedback modeling methods.}
\label{tab:reward}
\end{table*}

%% file: table/catagory.tex
\begin{table*}[t]
\centering

%\vspace{\baselineskip}
%\footnotesize
\resizebox{\textwidth}{!}{
\begin{tabular}{lllllc}
\toprule
Category & sub-category & Representative Methods & Tasks & Verifier/Critic & Train-free  \\

\midrule
\multirow{5}*{Repeat Sampling} & \multirow{2}*{Majority voting} & CoT-SC~\citep{wang2023selfconsistency} & Math, QA & self-consistency & $\checkmark$ \\
& & PROVE~\citep{toh2024votescountprogramsverifiers} & Math & compiler & $\checkmark$ \\
\cmidrule(r){2-6}
& \multirow{3}*{Best-of-N} & \citet{cobbe2021trainingverifierssolvemath} & Math & ORM & \ding{55}\\
& & DiVeRSe~\citep{li-etal-2023-making} & Math & PRM & \ding{55}\\
& & Knockout~\citep{liu2025pairwise} & Math & critic & $\checkmark$ \\
\midrule
\multirow{12}*{Self-correction} & \multirow{2}*{Human feedback} & NL-EDIT~\citep{elgohary-etal-2021-nl} & Semantic parsing & Human & \ding{55}\\
& & FBNET~\citep{tandon-etal-2022-learning} & Code & Human & \ding{55}\\
\cmidrule(r){2-6}
& \multirow{3}*{External tools} & DrRepair~\citep{Yasunaga20DrRepair} & Code & compiler & \ding{55}\\
& & Self-debug~\citep{chen2024teaching} & Code & compiler & $\checkmark$ \\
& & CRITIC~\citep{gou2024critic} & Math, QA, Detoxifying & text-to-text APIs & $\checkmark$\\
\cmidrule(r){2-6}
& \multirow{4}*{External models} & REFINER~\citep{paul-etal-2024-refiner} & Math, Reason & critic model & \ding{55} \\
& & Shepherd~\citep{wang2023shepherdcriticlanguagemodel} & QA & critic model & \ding{55}\\
& & Multiagent Debate~\citep{du2023improvingfactualityreasoninglanguage} & Math, Reason & multi-agent debate & $\checkmark$ \\
& & MAD~\citep{liang2024encouragingdivergentthinkinglarge} & Translation, Math & multi-agent debate & $\checkmark$ \\
%& & sparse MAD~(\citeyear{li2024improvingmultiagentdebatesparse}) \\
%& & ChainLM~(\citeyear{cheng-etal-2024-chainlm}) \\
\cmidrule(r){2-6}
& \multirow{3}*{Intrinsic feedback} & Self-Refine~\citep{madaan2023selfrefine} & Math, Code, Controlled generation & self-critique & $\checkmark$\\
%& & CoVe~(\citeyear{dhuliawala-etal-2024-chain}) & QA & self-critique \\
& & Reflexion~\citep{shinn2023reflexionlanguageagentsverbal} & QA & self-critique & $\checkmark$ \\
%& & Self-Critique~(\citeyear{saunders2022selfcritiquingmodelsassistinghuman}) & Summarization & self-critique & \\
& & RCI~\citep{kim2023languagemodelssolvecomputer} & Code, QA & self-critique & $\checkmark$ \\
\midrule
\multirow{6}*{Tree Search} & \multirow{2}*{Uninformed search} & ToT~\citep{NEURIPS2023_271db992} & Planing,  Creative writing & self-critique & $\checkmark$\\
& & \citet{xie2023selfevaluation} & Math & self-critique & $\checkmark$\\
\cmidrule(r){2-6}
& \multirow{4}*{Heuristic search} & RAP~\citep{hao-etal-2023-reasoning} & Planing, Math, Logical & self-critique & $\checkmark$\\
& & TS-LLM~\citep{feng2024alphazeroliketreesearchguidelarge} & Planing, Math, Logical & ORM & \ding{55} \\
%& & MCTS-DPO~(\citeyear{xie2024montecarlotreesearch}) \\
& & rStar~\citep{qi2024mutualreasoningmakessmaller} & Math, QA & multi-agent consistency & $\checkmark$\\
& & ReST-MCTS*~\citep{zhang2024restmctsllmselftrainingprocess} & Math, QA & PRM & \ding{55} \\
\bottomrule
\end{tabular}
}
\caption{Overview of search strategies.}
\label{tab:catagory}
\end{table*}

%% file: draft/5-advanced.tex
\section{Advanced Topics}
\label{advanced}
\subsection{Generalizable System-2 Model}
%泛化性体现在两个方面，一是对不同推理任务的泛化，二是真正的可泛化的推理能力，而不是记忆
Currently, most LRMs exhibit strong reasoning performance in specific domains such as math and code, but they struggle to generalize to cross-domain, cross-lingual, or general tasks~\citep{son2025linguisticgeneralizabilitytesttimescaling,zhao2025tradeoffslargereasoningmodels}. 
On the one hand, as the foundation of current System-2 models, CoT shows little effectiveness in non-symbolic reasoning tasks~\citep{sprague2024cotcotchainofthoughthelps}. On the other hand, verifiers and critics have limited generalization capabilities, making it difficult to provide effective guidance to the reasoning model~\citep{levine2023baseline,kim2024evaluating,chen2024odin}.
%The key to addressing this issue lies in enhancing the generalization ability of verifiers or critics.
For the former limitation, integrating external tools and multi-agent feedback is a promising direction~\citep{nathani-etal-2023-maf,lan2024training}. 
Many works aim to enable reasoning models to learn to perform retrieval or tool use during the reasoning process.
Search-o1~\citep{li2025searcho1agenticsearchenhancedlarge} triggers external search tools by generating special tokens that represent search actions, and Search-R1~\citep{jin2025searchr1trainingllmsreason} further enhances the interaction between the LLM and search tools through RL.
Deep Research~\citep{deepresearch,googledeepresearch,zheng2025deepresearcherscalingdeepresearch,li2025webthinkerempoweringlargereasoning} integrates various tools such as web search and code execution, demonstrating distinguished performance on general tasks like information synthesis, report writing, and complex code generation.
For the latter challenge, some works utilize multi-objective training~\citep{wang2024interpretable}, model ensemble~\citep{lin-etal-2024-dogerm}, soft reward~\citep{su2025crossingrewardbridgeexpanding} or regularization constraints~\citep{yang2024regularizing,jia2024generalizing} to make verifiers more generalizable.
Compared to discriminative verifiers, generative critics inherit the generalization ability of LLMs and can leverage test-time compute to realize greater potential~\citep{kim2025scalingevaluationtimecomputereasoning}.
GenPRM~\citep{zhao2025genprmscalingtesttimecompute} and GRM~\citep{liu2025inferencetimescalinggeneralistreward} leverage repeated sampling for test-time compute, and GRM trains a Meta RM to evaluate the quality of the critic's outputs.
RM-R1~\citep{chen2025rmr1rewardmodelingreasoning} and JudgeLRM~\citep{chen2025judgelrmlargereasoningmodels} use RL training to enable the critic with self-correction ability.

%Generalization reflects whether LRMs truly have reasoning ability rather than merely memorizing specific patterns. Although LRMs perform well on several math reasoning benchmarks, extensive empirical studies show that current System-2 models still lack genuine mathematical reasoning capability.

%Nevertheless, there is still significant room for improvement in the generalization of the verifier.
Additionally, weak-to-strong generalization~\citep{burns2023weaktostronggeneralizationelicitingstrong} is a topic worth further exploration. People are no longer satisfied with solving mathematical problems with standard answers; they hope System-2 models can assist in scientific discovery and the proofs of mathematical conjectures. In such cases, even human experts struggle to provide accurate feedback, while weak-to-strong generalization offers a promising direction to address this issue~\citep{tang2025enablingscalableoversightselfevolving}.

%\textcolor{red}{math robust \citep{yan2025recitationreasoningcuttingedgelanguage}}

\subsection{Multimodal Test-time Compute}
%
%However, test-time compute methods in System-2 thinking remain limited to text modalities. 
Visual, audio, video and other modalities are crucial for models understanding and interaction with the world. 
Like LLMs, large multimodal models (LMMs) also face challenges in scaling training compute. Thus, enabling efficient test-time scaling for multimodal tasks has become an advanced research topic.
In System-1 thinking, TTA has been successfully applied to LMMs, improving performance in tasks such as zero-shot image classification, image-text retrieval, and image captioning~\citep{zhao2024testtime}.
%To achieve cognitive intelligence, System-2 models must be able to fully integrate multimodal information for reasoning.
For multimodal reasoning tasks, the exploration of multimodal CoT~\citep{zhang2024multimodal,wu2024minds,mondal2024kam,lee-etal-2024-multimodal,gao2024interleavedmodalchainofthought,zhang2024improvevisionlanguagemodel} and multimodal critics or verifiers~\citep{xiong2024llava,tu2025vilbenchsuitevisionlanguageprocess} open up the possibility of building multimodal System-2 models.
For example, IoT~\citep{zhou2024imageofthoughtpromptingvisualreasoning} and SketchPAD~\citep{hu2024visual} utilize visual tools to obtain critical visual information and integrate it into CoT to refine multimodal reasoning.
VisualPRM~\citep{wang2025visualprmeffectiveprocessreward} constructs process annotation data through Monte Carlo estimation and trains a lightweight PRM verifier; and cutting-edge LMMs such as GPT-4o and Qwen-VL can serve as critics.
Building on these foundations, repeated sampling~\citep{lin2025investigatinginferencetimescalingchain} and tree search~\citep{yao2024mulberryempoweringmllmo1like,dong2024progressivemultimodalreasoningactive} can effectively further improve LMMs' multimodal reasoning performance.
\citet{xu2024llavacotletvisionlanguage} divide the visual reasoning process into four stages: task summary, caption, reasoning, and answer conclusion. They propose a stage-level beam search method, which repeatedly samples at each stage and selects the best result for the next stage.
Audio-Reasoner~\citep{xie2025audioreasonerimprovingreasoningcapability} transfers the four-stage CoT framework to the audio modality.
%Nowadays, Qwen team has released the open-weight multimodal reasoning model QVQ~\citep{qvq2024}, OpenAI and Kimi~\citep{kimiteam2025kimik15scalingreinforcement} have released their multimodal reasoning products. 
%For example, incorporating more modalities like speech and video into reasoning tasks, applying successful methods such as reflection mechanisms and tree search~\citep{yao2024mulberryempoweringmllmo1like,dong2024progressivemultimodalreasoningactive} to multimodal reasoning, or aligning the multimodal reasoning process with human cognitive processes.
To equip LMMs with the self-correction ability, Deepseek-R1’s training recipe has also been practiced in LMMs~\citep{ma2025rethinkingrlscalingvision,chen2025sftrlearlyinvestigation,shen2025vlmr1stablegeneralizabler1style}.
OpenVLThinker~\citep{deng2025openvlthinkerearlyexplorationcomplex} and Vision-R1~\citep{huang2025visionr1incentivizingreasoningcapability} convert images into text captions, enabling text-only LRMs to incorporate visual information for generating CoTs. This serves as cold-start data for further reinforcement learning to train multimodal reasoning models.
LMM-R1~\citep{peng2025lmmr1empowering3blmms} demonstrates that the reasoning ability acquired through text-only reinforcement learning can serve as an effective cold start for multimodal reinforcement learning training. 

Besides understanding and reasoning tasks, multimodal generation tasks can also benefit from test-time compute.
Mainstream multimodal generative models are divided into diffusion-based~\citep{peebles2023scalable,yang2025cogvideox} and autoregressive-based~\citep{tian2024visual,xie2025showo} models.
While diffusion models lack explicit CoTs, they can expand the search space at test time by sampling various Gaussian noises~\citep{xie2025sana15efficientscaling}.
For video generation, due to the high computational cost of repeatedly sampling full videos, \citet{cong2025testtimescalingimproveworld} use PRM to verify frames individually and apply beam search to prune low-scoring frames. \citet{liu2025videot1testtimescalingvideo} propose frame tree search, using critics to evaluate videos at early, middle, and late stages with different criteria, retaining only the top-k scoring frames at each step.
Autoregressive models treat intermediate generated images as reasoning steps. \citet{guo2025generateimagescotlets} propose a Potential Assessment Reward Model (PARM) to evaluate the quality potential of these intermediate images and introduce a reflection mechanism that enables the generative model to self-correct low-quality images.

\subsection{Efficient Test-time Compute}
The successful application of test-time compute shows that sacrificing reasoning efficiency can lead to better reasoning performance. 
However, researchers continue to seek a balance between performance and efficiency, aiming to achieve optimal performance under a fixed reasoning latency budget. 
This requires adaptively allocating computational resources for each sample.
\citet{damani2024learninghardthinkinputadaptive} train a lightweight module to predict the difficulty of a question, and allocate computational resources according to its difficulty.
\citet{zhang2024scalingllminferenceoptimized} further extend the allocation targets to more hyperparameters. %such as the model and temperature.
\citet{chen2025think23overthinkingo1like} and \citet{wang2025thoughtsplaceunderthinkingo1like} systematically evaluate the over-thinking and under-thinking phenomena in LRMs, where the former leads models to overcomplicate simple problems, and the latter causes frequent switching of reasoning paths on difficult problems, thereby reducing reasoning efficiency.
These phenomena even cause LRM to perform poorly when dealing with simple problems~\citep{zhang2025s1benchsimplebenchmarkevaluating}.
% O1-Pruner~\citep{luo2025o1prunerlengthharmonizingfinetuningo1like} and \citep{arora2025traininglanguagemodelsreason} propose the length-penalty reward to shorten reasoning processes while maintaining accuracy.
There are still many open questions worth exploring, such as how to integrate inference acceleration strategies, e.g. model compression~\citep{li-etal-2024-mode,huang2024o1,li2025quantization}, token pruning~\citep{fu2024lazyllm,zhang2024cut}, and speculative decoding~\citep{leviathan2023fast,xia2024unlocking} with test-time compute, and how to allocate optimal reasoning budget according to problem difficulty~\citep{wang-etal-2024-reasoning-token,han2024tokenbudgetawarellmreasoning,cheng2024compressedchainthoughtefficient}.

%distill:
%\citep{yu2024distilling21}
%preference optimization:
%\citep{wu2024thinkingllmsgeneralinstruction}
Long CoT leads to inference latency and memory footprints of key-value cache.
Recently, numerous studies have explored various strategies to reduce the reasoning length.
Early work primarily focus on SFT models~\citep{yu2024distilling21,NEURIPS2024_504fa7e5,kang2025c3ot,xia2025tokenskipcontrollablechainofthoughtcompression,ma2025cotvalvelengthcompressiblechainofthoughttuning,munkhbat2025selftrainingelicitsconcisereasoning,cui2025stepwiseperplexityguidedrefinementefficient,yang2025thinkingoptimalscalingtesttimecompute} using variable-length CoT data.
In contrast, mainstream approaches incorporate length-based rewards into reinforcement learning (RL)~\citep{kimiteam2025kimik15scalingreinforcement,luo2025o1prunerlengthharmonizingfinetuningo1like,arora2025traininglanguagemodelsreason,yeo2025demystifyinglongchainofthoughtreasoning,yi2025shorterbetterguidingreasoningmodels,yuan2025efficientrltrainingreasoning} to encourage concise responses.
\citet{xiao2025fastslowthinkinglargevisionlanguage} measure the complexity of vision-language reasoning tasks and design a complexity-aware curriculum learning and GRPO algorithm. They organize training samples from hard to easy and dynamically adjust the reward function and KL term weights based on task complexity.
L1~\citep{aggarwal2025l1controllinglongreasoning} and Elastic Reasoning~\citep{xu2025scalablechainthoughtselastic} can enable precise control over response length based on a given token budget.
Further, SelfBudgeter~\citep{li2025selfbudgeteradaptivetokenallocation} can autonomously estimate the optimal token budget before generation, and allow users to choose whether to wait for the full response or terminate early during the generation process.
However, both SFT and RL require substantial computational resources.
Several studies have instead explored training-free approaches for efficient reasoning, including prompting~\citep{han2024tokenbudgetawarellmreasoning,xu2025chaindraftthinkingfaster,aytes2025sketchofthoughtefficientllmreasoning,lee2025llmscompresschainofthoughttoken}, model merging~\citep{wu2025unlockingefficientlongtoshortllm,kimiteam2025kimik15scalingreinforcement} and early exit~\citep{fu2025reasoningdynasor,ma2025reasoningmodelseffectivethinking,yang2025dynamicearlyexitreasoning}.
Notably, DEER~\citep{yang2025dynamicearlyexitreasoning} makes early-exit decisions during CoT generation based on the confidence of intermediate answers, effectively reducing reasoning overhead.
Subsequently, ~\citet{dai2025sgrpoearlyexitreinforcement} introduces a new RL paradigm that encourages the model to incentivize early thinking termination when appropriate.
Another promising approach to achieving efficiency is hybrid reasoning, which dynamically switches between concise responses and long-chain reasoning based on task complexity.
These methods~\citep{lou2025adacotparetooptimaladaptivechainofthought,luo2025adar1hybridcotbileveladaptive,jiang2025thinkneedlargehybridreasoning,fang2025thinklessllmlearnsthink,zhang2025adaptthinkreasoningmodelslearn} follow a similar pipeline: they start with SFT on a mixed dataset of long and short CoT samples as a cold start, and then refine the policy optimization algorithm to train the model to acquire hybrid reasoning capabilities.

% concise CoT: prompting~\citep{xu2025chaindraftthinkingfaster,aytes2025sketchofthoughtefficientllmreasoning}; controllable via parameter interpolation~\citep{ma2025cotvalvelengthcompressiblechainofthoughttuning}

% length control: s1~\citep{muennighoff2025s1simpletesttimescaling} 'wait' and 'Final Answer' ;L1~\citep{aggarwal2025l1controllinglongreasoning} rl

% LightThinker~\citep{zhang2025lightthinkerthinkingstepbystepcompression} compresses thought steps into discrete special tokens

% sft~\citep{cui2025stepwiseperplexityguidedrefinementefficient}
% \citet{munkhbat2025selftrainingelicitsconcisereasoning,yang2025thinkingoptimalscalingtesttimecompute} sft; self-generate concise CoT by BoN and few-shot demonstration
% Tokenskip~\citep{xia2025tokenskipcontrollablechainofthoughtcompression} prune unimportant tokens then sft; controllable prune ratio

% distill into efficient architecture: mamba and hybrid Mamba~\citep{paliotta2025thinkingslowfastscaling}, Distilled Models Achieve Competitive Accuracy Under Fixed test-time compute budget; repeated sampling

% model deployment~\citep{lin2025sleeptimecomputeinferencescaling}

%% file: draft/5-future.tex
\section{Future Directions}
\label{future}
%\subsection{Generalizable System-2 Model}

%PRM can tolerate a modest amount of distribution shift~\citep{lightman2024lets}

%\subsection{Multimodal Reasoning}

%\subsection{Efficiency and Performance Trade-off}

\subsection{Test-time Scaling Law}
Unlike training-time computation scaling, test-time compute still lacks a universal scaling law.
Some works have attempted to derive scaling laws for specific test-time compute strategies~\citep{wu2024inferencescalinglawsempirical,levi2024simple}.
\citet{brown2024largelanguagemonkeysscaling} demonstrate that the performance has an approximately log-linear relationship with repeated sampling times.
\citet{chen2024simpleprovablescalinglaw} models repeated sampling as a knockout tournament and league-style algorithm, proving theoretically that the failure probability of repeated sampling follows a power-law scaling.
%\citep{akyürek2024surprisingeffectivenesstesttimetraining}
\citet{snell2024scalingllmtesttimecompute} investigate the scaling laws of repeated sampling and self-correction, and propose the computing-optimal scaling strategy.
There are two major challenges to achieving a universal scaling law: first, current test-time compute strategies are various, each with different mechanisms to steer the model; thus, it lacks a universal framework for describing them; second, the performance of test-time compute is affected by a variety of factors, including the difficulty of samples, the accuracy of feedback signals, and decoding hyperparameters, and we need empirical studies to filter out the critical factors.

%\citet{wu2025lessunderstandingchainofthoughtlength} long cot scaling law; consider task difficulty

\subsection{Strategy Combination}
Different test-time compute strategies are suited to various tasks and scenarios, so combining multiple strategies is one way to achieve better System-2 thinking.
For example, Marco-o1~\citep{zhao2024marcoo1openreasoningmodels} combines the MCTS and self-correction, using MCTS to plan reasoning processes, and self-correction to improve the accuracy of each step.
TPO~\citep{li2025testtimepreferenceoptimizationonthefly} combines BoN sampling and self-correction.
Moreover, test-time adaptation strategies in System-1 models can also be combined with test-time reasoning strategies.
\citet{akyürek2024surprisingeffectivenesstesttimetraining} combine test-time training with repeated sampling. They further optimize the language modeling loss on test samples, then generate multiple candidate answers through data augmentation, and finally determine the answer by majority voting.
They demonstrate the potential of test-time training in reasoning tasks, surpassing the human average on the ARC challenge.
Therefore, we think that for LLM reasoning, it is crucial to focus not only on emerging test-time strategies but also on test-time adaptation methods. By effectively combining these strategies, we can develop System-2 models that achieve or surpass o1-level performance.
%\citet{wu2024exampleshighlevelautomatedreasoning}

%\subsection{Broad Application Scenarios}

\subsection{New Test-time Compute Paradigms}
Latent CoT reasoning has emerged as a promising paradigm for test-time compute.
By performing internal reasoning in latent space, it overcomes the limitations of conventional explicit CoT in terms of computational efficiency and abstract reasoning capability~\citep{goyal2024think,xu2025softcotsoftchainofthoughtefficient,shen2025codicompressingchainofthoughtcontinuous,xu2025softcottesttimescalingsoft}.
Coconut~\citep{hao2024traininglargelanguagemodels} performs reasoning in a continuous latent space by directly feeding the model’s final hidden state as the next-step input embedding, thereby giving rise to advanced reasoning patterns.
CCoT~\citep{cheng2024compressedchainthoughtefficient} introduces a framework that generates continuous and variable-length contemplation tokens as compressed representations of reasoning chains, enabling efficient and accurate reasoning.
LightThinker~\citep{zhang2025lightthinkerthinkingstepbystepcompression} improves LLM efficiency by compressing intermediate reasoning steps into compact gist tokens via specialized data construction and attention mask design, enabling the model to discard verbose chains while maintaining the ability to reason over condensed representations.
Latent CoT currently faces several challenges, including the difficulty of direct supervision, limited generalization, and concerns over interpretability~\citep{chen2025reasoninglanguagecomprehensivesurvey}.
This calls for further exploration of token-level strategies and internal mechanisms to enable more robust and transparent latent reasoning.

%loop-transformer

%% file: draft/7-benchmark.tex
\section{Benchmarks}
\label{app-benchmark}
%\subsection{Benchmarks}
%\input{table/benchmark}
\paragraph{Test-time Adaptation}
In System-1 models, distribution shifts include adversarial robustness, cross-domain and cross-lingual scenarios.
In the field of CV, ImageNet-C~\citep{hendrycks2018benchmarking}, ImageNet-R~\citep{hendrycks2021many}, ImageNet-Sketch~\citep{wang2019learning} are common datasets for TTA.
\citet{yu2023benchmarking} propose a benchmark to conduct a unified evaluation of TTA methods across different TTA settings and backbones on 5 image classification datasets.
For NLP tasks, TTA is primarily applied in QA and machine translation tasks, with commonly used datasets such as MLQA~\citep{lewis-etal-2020-mlqa}, XQuAD~\citep{artetxe-etal-2020-cross}, MRQA~\citep{fisch-etal-2019-mrqa} and CCMatrix~\citep{schwenk-etal-2021-ccmatrix}.

\paragraph{Feedback Modeling}
RewardBench~\citep{lambert2024rewardbenchevaluatingrewardmodels} collects 20.2k prompt-choice-rejection triplets covering tasks such as dialogue, reasoning, and safety. It evaluates the accuracy of reward models in distinguishing between chosen and rejected responses.
RM-Bench~\citep{liu2024rm} further evaluates the impact of response style on reward models.
RMB~\citep{zhou2024rmbcomprehensivelybenchmarkingreward} extends the evaluation to the more practical BoN setting, where reward models are required to select the best response from multiple candidates.
CriticBench~\citep{lin-etal-2024-criticbench} is specifically designed to evaluate a critic model's generation, critique, and correction capabilities.
For PRM, \citet{song2025prmbench} propose PRMBench, which evaluates PRMs whether they can identify the earliest incorrect reasoning step in math tasks.
ProcessBench~\citep{zheng2024processbench} provides a more fine-grained evaluation, including redundancy, soundness, and sensitivity.
%Socratic-PRMBench~\citep{li2025socraticprmbenchbenchmarkingprocessreward}
In addition, there are benchmarks for evaluating multimodal feedback modeling, such as VL-RewardBench~\citep{li2024vlrewardbenchchallengingbenchmarkvisionlanguage}, VCR-Bench~\citep{qi2025vcrbenchcomprehensiveevaluationframework} and MJ-Bench~\citep{chen2024mjbenchmultimodalrewardmodel}.

\paragraph{Test-time Reasoning}
Reasoning capability is the core of System-2 models, including mathematics, code, commonsense, planning, etc~\citep{zeng2024mr}.
\textit{Math reasoning} is one of the most compelling reasoning tasks.
With the advancements in LLM and test-time compute, the accuracy on some previously challenging benchmarks, like GSM8K~\citep{cobbe2021trainingverifierssolvemath} and MATH~\citep{hendrycks2021measuring}, have surpassed the 90\% mark.
Thus, more difficult college admissions exam~\citep{zhang2023evaluating,arora-etal-2023-llms,azerbayev2024llemma} and competition-level~\citep{gao2024omnimathuniversalolympiadlevel} math benchmarks have been proposed.
Some competition-level benchmarks are not limited to textual modalities in algebra, logic reasoning, and word problems. For instance, OlympiadBench~\citep{he-etal-2024-olympiadbench}, OlympicArena~\citep{huang2024olympicarenabenchmarkingmultidisciplinecognitive} and  AIME~\citep{parvez_zamil_gollam_rabby_2024} provide images for geometry problems, incorporating visual information to aid in problem-solving, while AlphaGeometry~\citep{trinh2024solving} employs symbolic rules for geometric proofs.
The most challenging benchmark currently is FrontierMath~\citep{glazer2024frontiermathbenchmarkevaluatingadvanced}, with problems crafted by mathematicians and covering major branches of modern mathematics. Even the most advanced o3 has not achieved 30\% accuracy.

\textit{Code} ability is a key aspect of LLM reasoning, with high practical value, covering code completion~\citep{ding2023crosscodeeval,zhang-etal-2023-repocoder,gong2024evaluation}, code reasoning~\citep{gu2024cruxeval,wei2025equibenchbenchmarkinglargelanguage}, and code generation~\citep{chen2021evaluating,austin2021program} tasks. Among these, code generation gains more attention.
HumanEval~\citep{chen2021evaluating} and MBPP~\citep{austin2021program} provide natural language descriptions of programming problems, requiring LLMs to generate corresponding Python code and use unit tests for evaluation.
MultiPL-E~\citep{cassano2022multipl} extend them to 18 program languages.
EvalPlus~\citep{liu2024your} automatically augments test cases to assess the robustness of the generated code.
Recently, some studies collect benchmarks from open-source projects, which are closed to realistic applications and more challenging due to complex function calls, such as DS-1000~\citep{lai2023ds}, CoderEval~\citep{yu2024codereval}, EvoCodeBench~\citep{li2024evocodebench} and BigCodeBench~\citep{zhuo2025bigcodebench}.

\textit{Commonsense reasoning} requires LLMs to possess both commonsense and reasoning abilities.
%Early benchmarks~\citep{zellers-etal-2019-hellaswag,talmor-etal-2019-commonsenseqa,sakaguchi2021winogrande,bisk2020piqa} focus on evaluating LLMs' commonsense ability.
StrategyQA~\citep{geva-etal-2021-aristotle} collects complex and subtle multi-hop reasoning questions.
MMLU~\citep{hendrycks2021measuring} and MMLU-Pro~\citep{wang2024mmlupro} cover commonsense reasoning questions across various domains, including STEM, the humanities, the social sciences, etc.
\textit{Planning} aims to enable LLMs to take optimal actions based on the current state and environment to complete tasks. Current planning benchmarks primarily focus on synthetic tasks, such as Blocksworld~\citep{valmeekam2023planbench}, Crosswords, and Game-of-24~\citep{NEURIPS2023_271db992}.

%logic reasoning: MASTERMINDEVAL

% \subsection{Projects}
% \paragraph{OpenR}\citep{wang2024openr}\footnote{https://github.com/openreasoner/openr} is an open-source test-time reasoning framework that integrates various test-time compute strategies, PRM training, and improvement training. It currently supports beam search, BoN, MCTS, and rStar, and implements popular online reinforcement learning algorithms like APPO, GRPO, and TPPO.

% \paragraph{RLHFlow}\citep{dong2024rlhf} offers a comprehensive framework for reward modeling\footnote{https://github.com/RLHFlow/RLHF-Reward-Modeling} and online RLHF training\footnote{https://github.com/RLHFlow/Online-RLHF}. Its standout feature is the integration of various reward model training methods, including the vanilla preference reward model, multi-objective reward models, PRM, etc.

% \paragraph{OpenRLHF}\citep{hu2024openrlhfeasytousescalablehighperformance}\footnote{https://github.com/OpenRLHF/OpenRLHF} also integrates reward modeling and RLHF training but focuses more on the efficient implementation of reinforcement learning algorithms and training tricks. Its strength lies in the integration of distributed training and efficient fine-tuning, enabling users to easily train large language models with more than 70B parameters.

%% file: draft/6-conclusion.tex
\section{Conclusion}
In this paper, we conduct a comprehensive survey of existing works on test-time compute.
We introduce various test-time compute methods in System-1 and System-2 models, and look forward to future directions for this field. 
%We believe that test-time computing is a promising path toward cognitive intelligence, and hope this paper will promote further research in this area.
We believe test-time compute can help models handle complex real-world distributions and tasks better, making it a promising path for advancing LLMs toward cognitive intelligence. 
We hope this paper will promote further research.